\documentclass[letterpaper,10pt,conference]{ieeeconf}
\pdfoutput=1
\usepackage[utf8]{inputenc}
\usepackage{amsmath,xcolor}
\usepackage{graphicx}
\usepackage{epstopdf}
\usepackage{amssymb}
\usepackage{amsfonts}
\usepackage{url,comment}
\usepackage{amssymb}
\usepackage{color}
\usepackage{booktabs}
\usepackage{multicol}
\usepackage{multirow}
\usepackage{cite}
\usepackage{colortbl}
\usepackage{amsfonts}
\usepackage{subfigure}
\usepackage{algorithm, algorithmic}
\usepackage{todonotes}
\usepackage{changes}

\definechangesauthor[name=Longsheng, color=purple]{LJ} 
\definechangesauthor[name=Zhaojian, color=purple]{ZL} 
\definechangesauthor[name=Dong, color=blue]{DC}

\newcommand{\dcal}{\mathcal{D}}

\usepackage{nomencl}

\makenomenclature

\usepackage{makeidx}
\makeindex

\usepackage{tikz}
\usetikzlibrary{shapes,backgrounds,calc}

\makeatletter
\tikzset{circle split part fill/.style  args={#1,#2}{%
 alias=tmp@name, 
  postaction={%
    insert path={
     \pgfextra{%
     \pgfpointdiff{\pgfpointanchor{\pgf@node@name}{center}}%
                  {\pgfpointanchor{\pgf@node@name}{east}}%
     \pgfmathsetmacro\insiderad{\pgf@x}
      \fill[#1] (\pgf@node@name.base) ([xshift=-\pgflinewidth]\pgf@node@name.east) arc
                          (0:180:\insiderad-\pgflinewidth)--cycle;
      \fill[#2] (\pgf@node@name.base) ([xshift=\pgflinewidth]\pgf@node@name.west)  arc
                           (180:360:\insiderad-\pgflinewidth)--cycle;            
         }}}}}
 \makeatother


\title{\LARGE \bf Autonomous Driving using Safe Reinforcement Learning by Incorporating a Regret-based Human Lane-Changing Decision Model}
\author{Dong Chen$^*{^1}$, Longsheng Jiang$^*{^2}$, Yue Wang$^2$, Zhaojian Li$^1$
\thanks{$^1$Dong Chen and Zhaojian Li are with the Department of Mechanical Engineering, Michigan State University, East Lansing, MI 48824, USA. Email: {\tt\small \{chendon9,lizhaoj1\}@egr.msu.edu.}}
\thanks{$^2$Longsheng Jiang and Yue Wang are with the Department of Mechanical Engineering, Clemson University, Clemson, SC 29634, USA. Email: {\tt\small \{longshj,yue6\}@g.clemson.edu.}}
\thanks{$*$Both authors contributed equally to this work.}}

\IEEEoverridecommandlockouts

\begin{document}

\maketitle
\begin{abstract} 

 It is expected that many human drivers will still prefer to drive themselves even if the self-driving technologies are ready. Therefore,  human-driven vehicles and autonomous vehicles (AVs) will coexist in a mixed traffic for a long time. To enable AVs to safely and efficiently maneuver in this mixed traffic, it is critical that the AVs can understand how humans cope with risks and make driving-related decisions.
On the other hand, the driving environment is highly dynamic and ever-changing, and it is thus difficult to enumerate all the scenarios and hard-code the controllers. 
To face up these challenges, in this work, we incorporate a human decision-making model in reinforcement learning to control AVs for safe and efficient operations.
Specifically, we adapt regret theory to describe a human driver's lane-changing behavior, and fit the personalized models to individual drivers for predicting their lane-changing decisions.
The predicted decisions are incorporated in the safety constraints for reinforcement learning in training and in implementation.
We then use an extended version of double deep Q-network (DDQN) to train our AV controller within the safety set. 
By doing so, the amount of collisions in training is reduced to zero, while the training accuracy is not impinged.

\end{abstract}

\begin{keywords}
Safe Reinforcement Learning, Human Lane-changing Decisions, Regret Theory, DDQN 
\end{keywords}

\section{INTRODUCTION}\label{sec:1}

Autonomous driving has attracted significant research interest in the past two decades as it offers the potential to release drivers from exhausting driving. 
%
While great progresses have been made in the field of perception, path planning, and controls, high-level decision-making remains a big challenge due to the involvement of complex, cluttered environment and the dynamic, uncertain behaviors of other traffic users. Some recent works have been applying reinforcement learning (RL) methods to autonomous driving and promising performance \cite{wang2018reinforcement} has been reported.
RL-based methods can learn the decision-making and driving behaviors which are hard, if not infeasible, for traditional rule-based designs, and often with much less human effort.
%


However, it is reported in \cite{DBLP:journals/corr/abs-1904-00035}  that when using RL-based methods lots of collisions happen before the agent starts to behave properly.
Although these collisions are not prohibitive in simulations, in practice, the RL-based driving algorithms must be trained, adapted, and tested in real traffic, where  collisions can cause disastrous consequences. 
In implementation phase, autonomous vehicles (AVs) using the trained RL algorithms may not behave safely in unseen driving environment, and the trained models may choose unsafe actions due to function approximation \cite{silver2016mastering} used in most RL algorithms, therefore, making the implementation risky.

To tackle the aforementioned issues and provide safety guarantees in RL, the idea of safe reinforcement learning (SafeRL) \cite{li2018safe} has been proposed where safety supervisors are deployed to ensure safe explorations and exploitation for the RL agents. 
Nageshrao et. al. \cite{DBLP:journals/corr/abs-1904-00035} incorporate a short-horizon safety check in the RL-based method. 
The supervisor replaces  identified risky actions with safe ones during training and implementation. 
The collisions were significantly reduced. 
Wang et al. \cite{wang2019lane} developed a rule-based decision-making framework for lane-changing.
The framework examines the trajectories prescribed by the controller and changes the actions resulting in collisions. In \cite{li2018safe}, a dynamics-enabled safe RL framework is developed to train a fuel-efficient adaptive cruise control policy without collisions.

It is worth noting that when supervising the learning process to avoid collisions, oversimplified, non-interactive environment vehicles are considered in the aforementioned safeRL studies.
To enable AVs safely interacting with manual-driven vehicles (MVs), it is crucial to understand and characterize how human drivers make driving-related decisions when they interact with other road users.
%
Extant models of human driving behaviors are either data-driven \cite{macadam1996application} or motivational \cite{michon1985critical}.
While data-driven methods lack explaniability \cite{macadam2003understanding}, motivational models ignore the usefulness of data and fail in generating testable predictions \cite{ranney1994models}.


Risks in driving have two dimensions: harm (costs) and probabilities. 
Human drivers most times manage these risks well. 
Regret theory \cite{loomes1982regret} in behavioral economics is a good candidate for modeling human decision-making under risks. 
It emphasizes the regret effect: comparing the costs caused by different actions induces anticipated
regretful emotion, which in turn biases the comparison.
Two other psychological effects are also important.
The probability weighting effect claims that human brains process the probability they see nonlinearly \cite{hsu2009neural}.
The range effect asserts that when evaluating the costs, not only values of the costs themselves but the range of the costs contribute to their evaluated psychological goodness or badness (utilities) \cite{kontek2017range}. 

An extended regret decision model was developed in our prior work \cite{jiang2019human} for describing the three psychological effects quantitatively.
In this paper, the parametric regret decision model is further evolved such that the abstract harm and probabilities can be described with the physical terms like speeds and distances.
Similar to motivational models, the regret decision model is explainable.
Also, after estimating the parameters using drivers' data, the model can predict a driver's decisions, hence the movement of the MV.
These predictions can facilitate the training safety of AVs.


In this paper, we integrate a regret-based human decision-making model into a safeRL framework to enable AVs to learn a safe and efficient policy in a challenging scenario. We  show that the regret-based decision model is able  to estimate reasonably accurate driver-related decisions in this new application.
We design a hierarchical learning structure that includes  a RL-based decision-making agent with an extended double Q-network (DDQN) \cite{van2016deep} and a safety supervisor that uses the driver lane-changing model to identify unsafe actions. 
We exploit both  safe and unsafe experiences for training to improve the learning efficiency. The efficacy of the proposed framework is demonstrated in CARLA~\cite{Dosovitskiy17}. 





The reminder of the paper is organized as follows. Section \ref{sec:2} formulates the research problem.
In Section \ref{sec:3}, we show how to build the regret-based human lane-changing model. 
Our proposed safeRL algorithm is depicted in Section \ref{sec:4}. 
Experiments, results and discussion are shown in Section \ref{sec:6}. 
Conclusion and future works are discussed in Section \ref{sec:7}.

\section{Problem Formulation}\label{sec:2}


\subsection{Traffic Scenario} 
\begin{figure}[!ht]
  \centering
  \includegraphics[width=0.9\linewidth]{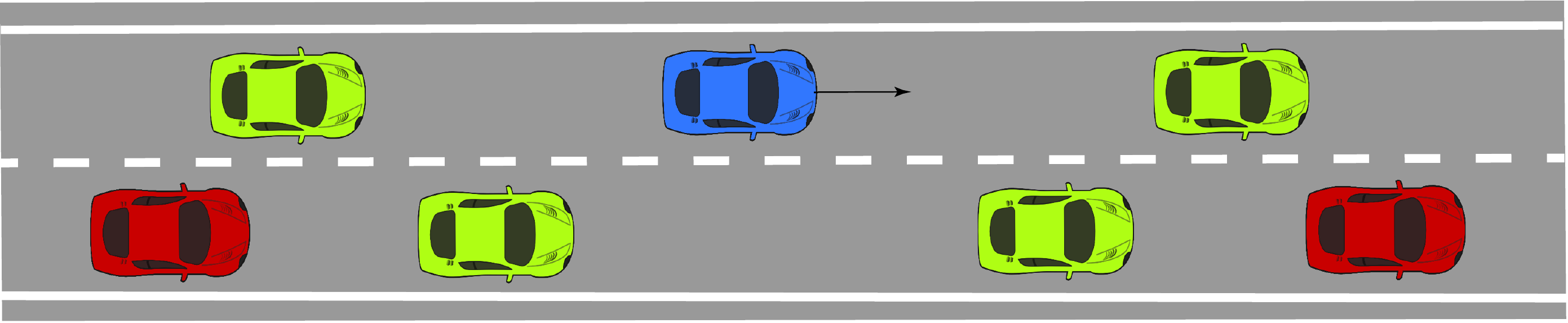}
  \caption{Two-lane traffic scenario}\label{fig:AV_traffic}
\end{figure}

In a two-lane highway scenario, as shown in Fig.~\ref{fig:AV_traffic}, the ego vehicle (blue, AV) is surrounded by environment vehicles (green) which are MVs.
Vehicles (red) far away from the ego vehicle are not considered. 
All the vehicles are heading longitudinally.
They currently have different speeds and they all want to run at their desired speeds.
To achieve that goal, they need to change lanes whenever necessary while maintaining safe.
The AV is controlled by a RL-based intelligent agent. 
The agent learns how to drive, including longitudinal speed control, lane-changing strategy, etc, from interacting with the environment vehicles. 
The AV is equipped with sensing systems. 
Because the statue quo is that there is no communication between vehicles, we assume at least no communication exists involving MVs.
Each MV makes lane-changing decisions independently. 

\subsection{Training with Conventional RL}
When using conventional RL for training an AV, state $s_t$ represents traffic scenarios, including the velocities, locations of the vehicles, and the agent applies action $a_t$ to navigate.
Given a reward function $r_{s_t, a_t}$, the optimal policy $\pi^{*} (s_t)$ is to maximize the expected cumulative future rewards:
\begin{equation}
R \triangleq \mathbf{E}_{\pi} \left[\sum_{t=0}^{\infty} \gamma ^t r_{s_t, a_t}\right]    
\end{equation}
where scalar $\gamma$ is the discount factor.

Q-learning \cite{watkins1989learning} is a model-free method which works well on discrete action space.
It uses a function approximator  $Q_{\phi}(s_t,a_t)$ to approximate the Q function, where the optimal Q value is defined by ${Q ^ {*}}(s_t,a_t)$, and  $\phi$ are the parameters of the approximator.
The action is chosen by $a_t = \max_{a'_t} Q_{\phi}(s_t, a'_t)$. 

Deep Q network (DQN) \cite{volodymyr2015human, mnih2013playing} uses the deep neural networks to approximate the Q function.
It stores the explored experiences into a replay buffer and samples $K$ experiences each time to update the parameters,
\begin{equation}\label{eqn:update_phi}
{\phi} 	\leftarrow  {\phi} + \alpha \big(Y^Q_{\text{target}} - Q_{\phi}(s_{t}, a_{t}) {\big)}\nabla_{\phi} Q_{\phi}(s_{t}, a_{t}) 
\end{equation}
where $\alpha$ is the learning rate. 
The target $Y^Q_{\text{target}}$ is defined as 
\begin{equation}
Y^Q_{\text{target}} \triangleq r_{t+1} + \gamma \max_{a'} Q_{{\phi_N}}(s_{t+1}, a')  
\end{equation}


The parameters $\phi_N$ of the target network are updated only every $N$ steps by $\phi_N \leftarrow \phi$ and keep fixed at other steps.

\subsection{The Drawback of Conventional RL}
There is one drawback with conventional RL.
As we tested it on autonomous driving in two-lane traffic (Section V), we found that when using conventional RL, about 14.5\% training epochs ended with collisions.
Even after the policy converged, still 3.46\% of the trials caused collisions.
%
Collisions mainly came from the fact that the AV cannot estimate the intentions of MVs.
Especially for lane-changing, collisions can happen either because
\begin{itemize}
    \item the ego vehicle changes its lane but collides with the vehicles already in that lane;
    \item or, an environment vehicle suddenly changes to the ego vehicle's lane and the ego vehicle cannot react timely.  
\end{itemize}

Collisions cause training unstable as the RL algorithm needs to reset the simulation whenever collisions happen.
Real deployment of AVs will also not tolerate any collision. 

\begin{figure}[!ht]
  \centering
  \includegraphics[width=0.47\textwidth]{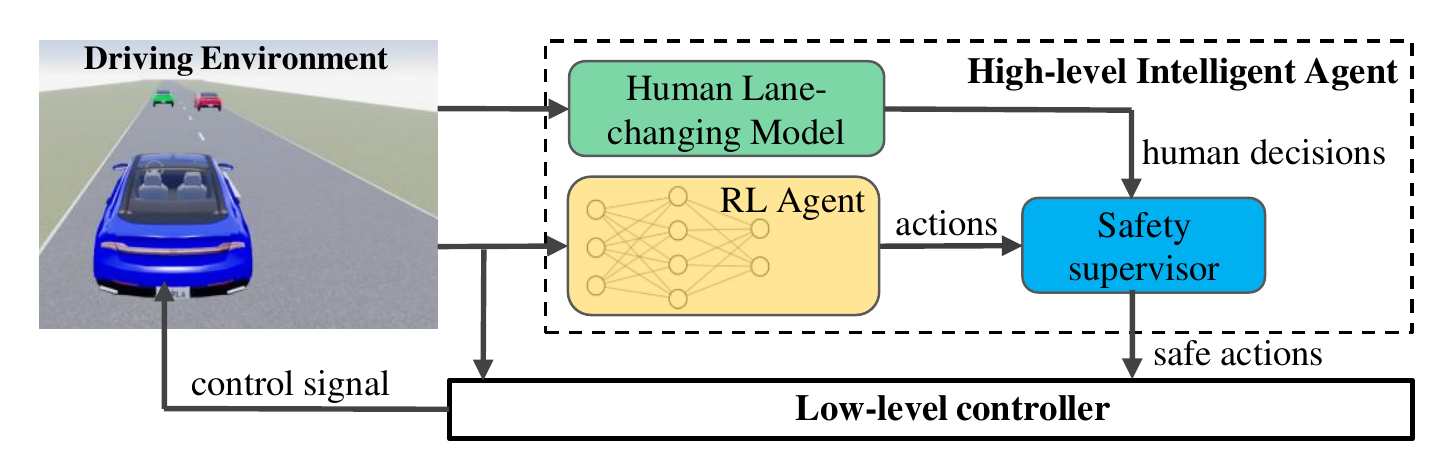}
  \caption{Framework of SafeRL system.}\label{SafeRL_system}
\end{figure}

To solve the problem, we present a framework for RL to incorporate a safety supervisor which uses human lane-changing decision model for making predictions.
The architecture is given by Fig.~\ref{SafeRL_system}. 

We adopt a hierarchical structure. 
A high-level intelligent agent in the ego vehicle 
receives states of the traffic in the driving environment. 
Inside the agent, based on the current traffic scenario, the human-lane changing model (Section \ref{sec:3}) estimates the lane-changing decisions that will be taken by the MVs. 
In parallel, the RL agent determines optimal actions based on the states (Section \ref{sec:4}).
The actions from the RL agent are supervised by the safety supervisor: The supervisor uses the predicted human decisions to check if an action is safe.
The unsafe actions are replaced with the safe actions. 
The safe actions go to the low-level controller for navigating the ego vehicle in the driving environment.

\section{Regret-based human lane-changing model}\label{sec:3}

In this section, we will present how to build a human lane-changing decision model.
Within the RL framework, the AV needs to predict actions of MVs, which are the results of the drivers' decisions.
One of the most safety critical decisions in the two-lane traffic in Fig. \ref{fig:AV_traffic} is whether a driver intends to change lanes. 
When a driver can drive at the desired speed, or faster than in the neighboring lane, the decision is straightforward: staying in the current lane. 
However, when traffic in the current lane is slower, the decision-making becomes hard to predict. We will focus on this situation.

\subsection{Human Decision-making in Two-lane Traffic}

\begin{figure}[!ht]
  \centering
  \includegraphics[width=0.9\linewidth]{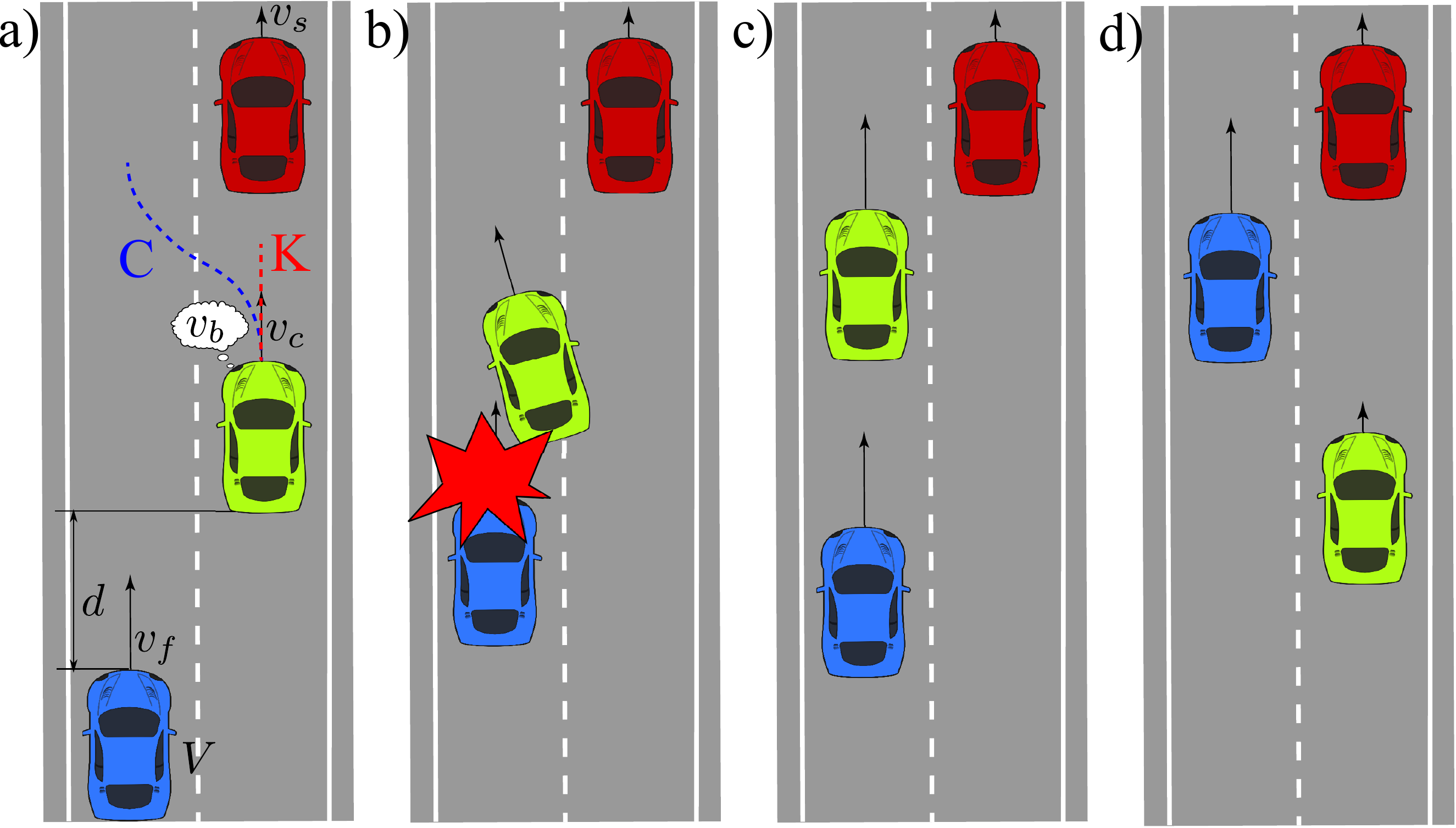}  
  \caption{The MV (green) in the traffic in (a) can either change lanes or keep lanes. If it changes lanes, it may collide with the approaching vehicle (b) or safely merge in (c); otherwise, it has to slow down (d).}
  \label{fig:TrafficScenario}
\end{figure}

From the perspective of the green MV (Fig. \ref{fig:TrafficScenario}a),
it runs at speed $v_c$ in the right lane. The driver has a best possible speed $v_b$ in mind, but the current lane is blocked by an environment vehicle (red) driving at a speed $v_s \leq v_b$.
In the other lane, there is a stream of traffic running at a faster speed $v_f \geq v_s$. 
The vehicle (blue), which approaches the MV longitudinally, has the size of volume $V$ and is currently at speed $v_f$. 
The gap between the approaching vehicle and the MV currently is $d \geq 0$. 
The speeds $v_s,\, v_c,\, v_f$, the distance $d$, and the volume $V$ 
can be observed by the MV and the approaching vehicle.
Speed $v_b$ seems to be private to the driver in the MV.
However, often it can be inferred by the approaching vehicle, because it usually either is the speed limit or the speed of the fast lane traffic. 
Although in Fig. \ref{fig:TrafficScenario}, the right lane is slow, because of symmetry, when the left lane is slow, similar situations can happen. 

The driver in the MV has two options. 
He/she can either make a lane change (option C) or keep the current lane and yield to the approaching vehicle (option K). 
If the driver decides to change lanes, there may be two possible outcomes. 
The MV may rear collide with the approaching vehicle, as shown in Fig. \ref{fig:TrafficScenario}b, or it may successfully merge in the new lane and run at its best speed $v_b$ as shown in Fig. \ref{fig:TrafficScenario}c.
The blocking vehicle is assumed not braking abruptly.
If the driver chooses to yield, the MV has to slow down to match the speed of the blocking vehicle, letting the approaching vehicle pass (Fig. \ref{fig:TrafficScenario}d).  
Which option will the driver take, and how is the decision made?


\subsection{Regret Decision Model}
The regret decision model deals with two options that are formulated in terms of harms (costs) and probabilities.
To use regret decision model on the lane-changing problem above, the two options, option C and option K, are expected to take the form of Table \ref{tb:twooption}.

\begin{table}[!h]
\renewcommand{\arraystretch}{1.3}
\caption{Option C and option K in terms of harm and probabilities}
\label{tb:twooption}
\centering
\begin{tabular}{c c c c c c c}
\multicolumn{3}{c}{\bf  Option C} & & \multicolumn{3}{c}{\bf  Option K}\\
\cmidrule{1-3} \cmidrule{5-7}
Cost:   &   $ 0$    &     $c_\text{collide}$ &  & Cost:   &  \multicolumn{2}{c} { $c_\text{slow}$}\\
Probability: & $p$ &      $1-p$&  & Probability:  & \multicolumn{2}{c} {$1$}\\
\cmidrule{1-3} \cmidrule{5-7}
\end{tabular}
\end{table}

Choosing option C can either lead to no harm (Fig. \ref{fig:TrafficScenario}c) with an objective and observable probability $p$, or lead to a cost of collision $c_\text{collide}$ (Fig. \ref{fig:TrafficScenario}b) with the probability $1-p$. 
Choosing option K is always safe.
But deceleration generates some cost $c_\text{slow}$ (Fig. \ref{fig:TrafficScenario}d), with probability 1. 

The regret decision model includes three components to address the three psychological effects, namely, the range effect, the regret effect, and the probability weighting effect. 
The range effect claims that the utility $u$ of a cost $c$ is relevant to a reference cost $|\bar{c}|$ that defines the overall severity of the decision-making problem. 
In the lane-changing problem, a collision is the most critical event that may happen.
Cost $c_\text{collide}$ thus draws drivers particular attention and is used as the reference to judge the situation. 
Hence, it is plausible that $|\bar{c}| \triangleq |c_\text{collide}|$.
The utilities are defined as the scaled costs \cite{kontek2017range}: $u \triangleq c/|c_\text{collide}|$. 

Since the anticipated regret emotion in the regret effect is triggered by the comparison of costs in different options, when using regret decision model, it is convenient to transform Table \ref{tb:twooption} to Table \ref{tb:twooption_comparison}. 

\begin{table}[h!]
\renewcommand{\arraystretch}{1.3}
\caption{Option C and option K represented in the comparative form}
\label{tb:twooption_comparison}
\centering
\begin{tabular}{c | c| c| c}
\hline
\multicolumn{2}{c|}{Events:} & No collision & Collision \\\hline
\multicolumn{2}{c|}{Joint probability:} & {$p$} & $1-p$ \\ \hline
\multirow{2}{*}{Cost:} &   {\bf Option C} & $0$ & $u_\text{collide}$ \\ \cline{2-4}
                       &   {\bf Option K} & $u_\text{slow}$ & $u_\text{slow}$ \\  \hline
\end{tabular}
\end{table}

Each column is a comparison between utilities. 
The probability in a column is the joint probability for both utilities in the comparison to happen. 
Since the influence of regret emotion is nonlinear, the regret effect is depicted as \cite{liao2017quantitative},
\begin{equation}\label{eqn:q}
    q(\Delta u) \triangleq \sigma_1 \sinh (\sigma_2 \Delta u) + \sigma_3 \Delta u,
\end{equation}
where $\Delta u$ is the difference between utilities, e.g., $u_\text{collide}- u_\text{slow}$ (column 4 in Table \ref{tb:twooption_comparison}).
The parameters $\sigma_1, \sigma_2, \sigma_3 \geq 0$ are specific to individuals and will be determined ( Section \ref{sec:6}).
The linear part, i.e., $\sigma_3 \Delta u$, represents the objective (so-called rational) evaluation of utility differences;
the nonlinear part, i.e., $\sigma_1 \sinh (\sigma_2 \Delta u)$, represents the influence of regret emotion triggered by cost differences. 

The probability weighting effect is described by another nonlinear function \cite{prelec1998probability}:
\begin{equation}
\label{eqn:w}
    w(p) \triangleq \exp (-\beta_1 (-\log(p))^{\beta_2}),
\end{equation}
where parameters $\beta_1, \beta_2 \geq 0$ are specific to individuals and will be determined (Section \ref{sec:6}). 
Comparing with the naive weighting $w(p)=p$, Eqn. (\ref{eqn:w}) either overweights or underweights objective probability $p$. 

To use the three components, $|\bar{c}|$, $q(\Delta u)$, and $w(p)$, to make a decision, the regret decision model calculates the net advantage of option C over option K:
\begin{equation}
\label{eqn:regret_decision_model}
    e_{ck} \triangleq w(p)\, q(0-u_\text{slow}) + (1-w(p))q(u_\text{collide}-u_\text{slow}).
\end{equation}
When $e_{ck}>0$, choose option C; otherwise choose option K. 

\subsection{Human Drivers' Lane-changing Decision-making}
The driver in the MV, however, is not readily presented with the values of objective probability $p$ and the objective costs $c_\text{collide}$ and $c_\text{slow}$ as in Table \ref{tb:twooption}. 
What he/she knows and can observe are the physical terms that define the traffic, including the speeds $v_s,\, v_b,\, v_c,\, v_f$, the distance $d$, and the volume $V$ of the approaching vehicle. 
To use the regret decision model, we should develop a bridge between the traffic defining terms and the costs and probabilities. 

Not all the drivers have experienced collisions, but all of them can perceive the threat of a potential collision. 
It is plausible to assume that the driver in the MV perceives the threat of the approaching vehicle to be proportional to its kinematic energy $\frac{1}{2}\rho V v_f^2$. 
This is because, as common sense, the vehicles of larger size are always perceived of greater threat, and the threat grows more than linearly when vehicle speed increases. 
Hence, we define 
\begin{equation}
 c_\text{collide}(v_f, V) \triangleq \lambda \frac{\rho V v_f^2}{2},   
\end{equation}
where parameter $\lambda<0$ is the subjective threat factor.  

We also speculate that the cost of slow-down $c_\text{slow}$ is the time loss and is defined as 
\begin{equation}\label{eqn:c_slow}
    c_\text{slow}(v_d, v_o) \triangleq \frac{d_a}{v_d}- \frac{d_a}{v_o} = \tau_a \left(1 - \frac{v_d}{v_o}\right).
\end{equation}
In Eqn. (\ref{eqn:c_slow}), we introduce two new variables: $v_d$ and $v_o$.
While the desired speed of the driver is $v_d$, choosing one option the MV will run at speed $v_o$.
Note, as discussed below, the desired speed $v_d$ may not necessarily equal the best speed $v_b$. 
Over some anticipated distance $d_a$, the different speeds generate the time loss $ c_\text{slow}(v_d, v_o)$.
One example in daily life is that when we are stuck in a traffic jam, we may feel grilled, because stopping at the spot means the time to reach our destinations is infinitely long. 
Again, the anticipated distance $d_a$ to travel is subjective.
It is defined as $d_a \triangleq \tau_a\, v_d$, where parameter $\tau_a \geq 0$ is the subjective anticipated impediment-free time. 

The arguments $v_d$ and $v_o$ in cost $c_\text{slow}(v_d, v_o)$ take different values when the driver is reasoning over the options and conditions. 
If option C was chosen and no collision could happen, the desired speed for the MV was $v_d =v_b$, and the MV could run at $v_o = v_b$. 
In this case, cost $c_\text{slow} = 0$. 
If no collision could happen for choosing option C but the driver was considering to choose option K, the desired speed was $v_d = v_b$, and the MV had to match the speed of the blocking vehicle with $v_o = v_s$. In this case, the slow-down cost became $c_\text{slow} = \tau_a - \frac{\tau_a v_b}{v_s}$.
If collision could happen when option C was taken, option K actually became the only feasible option: The driver had no choice but took option K.
The desired speed for the driver became $v_d = v_s$, and the MV ran at $v_o = v_s$. 
The slow-down cost is $c_\text{slow} = 0$. 
If collision could happen when option C was taken but the driver was considering to choose option C, this decision would realize the collision, causing cost $c_\text{collide}(v_f, V)$.
  
The objective probability $p$ to make a successful lane change is also hidden from the driver; 
what can be observed are the distance $d$ and the relative speed $v_f - v_c$.
The driver must estimate a probability $\hat{p}$ directly from the observed physical variables. 
We speculate that probability $\hat{p}$ strongly relates to the time-to-collision with the approaching vehicle $t_c$ that is defined as 
\begin{equation}
    t_c \triangleq
    \begin{cases}
          \frac{d}{v_f - v_c}, & v_c < v_f; \\
          \infty, & v_c \geq v_f.
    \end{cases}
\end{equation}
We further speculate that $t_c$ is compared against a subjective time constant $\tau_s$ which represents the duration for safely and comfortably changing lanes. 
The estimated probability $\hat{p}$ is
\begin{equation}
    \hat{p} \triangleq
    \begin{cases}
    \frac{t_c}{\tau_s}, & 0 \leq t_c \leq \tau_s; \\
    1,                  & t_c > \tau_s. 
    \end{cases}
\end{equation}

The definitions of costs and probabilities and the definition of utilities, $u = c /|c_\text{collide}|$, express Table \ref{tb:twooption_comparison} in terms of the observable physical variables. 
If we further assume all the vehicles in the traffic are of the same size, as in Fig. \ref{fig:TrafficScenario}, we can treat volume V as a constant parameter.
To reduce the total amount of parameters, we define $\eta_1 \triangleq -\frac{2\tau_b}{\lambda \rho V} \geq 0$. 
The two options faced by the driver are in Table \ref{tb:twooption_physical}.
Based on (6), The decision of the driver is modelled as
\begin{equation}\label{eqn:driver_decision_model}
    e_{ck} = w(\hat{p})\, q\Big(\eta_1\Big(\frac{v_b}{v_s v_f^2}-\frac{1}{v_f^2} \Big)\Big) + (1-w(\hat{p}))q(-1),
\end{equation}
where functions $w(\cdot)$ and $q(\cdot)$ are defined in Eqns. (\ref{eqn:w}) and (\ref{eqn:q}), respectively.
When $e_{ck}>0$, choose option C; choose option K, otherwise. 
\begin{table}[h!]
\renewcommand{\arraystretch}{1.3}
\caption{Option C and option K in terms of physical variables}
\label{tb:twooption_physical}
\centering
\begin{tabular}{c | c| c| c}
\hline
\multicolumn{2}{c|}{Events:} & No collision & Collision \\\hline
\multicolumn{2}{c|}{Joint probability:} & {$\hat{p}$} & $1-\hat{p}$ \\ \hline
\multirow{2}{*}{Cost:} &   {\bf Option C} & $0$ & $-1$ \\ \cline{2-4}
                       &   {\bf Option K} & $\eta_1\left( \frac{1}{v_f^2} -\frac{v_b}{v_s v_f^2} \right)$ & $0$ \\  \hline
\end{tabular}
\end{table}

\section{Safe RL algorithm}\label{sec:4}


In this section, we will focus on the development of a safe RL algorithm that integrates a safety supervisor.
First, we define the state representation, the action space and the reward function for the reinforcement learning agent.
Then, we demonstrate how to incorporate the human lane-changing model as safety supervisor into the RL algorithm.

\subsection{Reinforcement Learning Agent}
\subsubsection{States Representation}
 Methods like \cite{xu2017end} use mediated perception, by stacking multiple input images, and encode the images to low dimensional states for representing the world. 
 These methods not only need carefully designed encoders but also need to collect manually labeled data.
They are really time-consuming and hard to train.
 Recently, works \cite{Chen_2015_ICCV, li2017explicit} start to use the affordance indicators as the world representation.
 It shows advantages over the mediated perception: they only use a small number of key world indicators which are obtained from sensing systems, such as vehicles' locations on roads, relative distances and speeds between vehicles.

 In this work, we also use the affordance indicator method.
 For a two-lane road as shown in Fig.\ref{fig:AV_traffic}, assume we need the following indicators to represent the world.

\begin{itemize}

\item In front of the ego vehicle, the relative distances and the relative velocities of the vehicle in the right lane are $d_{fr}$ and $v_{fr}$; those of the vehicle in the left are $d_{fl}$ and $v_{fl}$.

\item To the rear of the ego vehicle, the relative distances and the relative velocities of the vehicle in the right lane are $d_{rr}$ and $v_{rr}$; those of the vehicle in the left are $d_{rl}$ and $v_{rl}$.

\end{itemize}

Besides the above 8 affordance indicators, we also include the lateral position $y$, longitudinal velocity $v_x$, steering angle $\theta$ and throttle value of the ego vehicle. 
A total of 12 affordance indicators are considered as the input to the RL agent.
For generalization consideration, we normalize all indicators to range $[-1, 1]$.

\subsubsection{Action Space}\label{sec:action_space}
Laterally, the ego vehicle can take two actions, turning left or right.
We assume the ego vehicle uses constant lateral speed $v_y \in \{-\bar{v}_y, 0, \bar{v}_y\}$ for lane changing. 
Longitudinally, there are three actions, decelerating, cruising or accelerating, i.e., $a_x \in \{-\bar{a}_x, 0, \bar{a}_x\}$.
The RL agent chooses one of the above actions each time. 
If the action is safe, it is sent to the low-level controller for generating control signals.

\subsubsection{Reward Function}
RL algorithms rely on reward functions to guide the agent to learn the desired policy. 
A reward/cost function penalizes the agent for choosing dangerous actions and rewards actions that bring efficiency, safety, and comfort.
Here we adopt a linear reward function as
\begin{equation}\label{eqn:reward_function}
r = w_s r_s + w_v r_v + w_c r_c + w_h  r_h,
\end{equation}
where $w_s$, $w_v$, $w_c$ and $w_h$ are weighting parameters for collision evaluation $r_s$, stable-speed evaluation $r_v$, lane-centering evaluation $r_c$, and headway evaluation $r_h$, respectively.
Safety is the most important criteria, so we choose $w_s \gg w_v, w_c, w_h$.

The various performance evaluations are defined as follows.
The collision evaluation is defined as a binary function:
\begin{equation}
    r_s \triangleq 
    \begin{cases}
          -1, & \text{Collision happened,} \\
          0, & \text{Otherwise}.
    \end{cases}
\end{equation}
We encourage the ego vehicle to run at a stable speed.  Hence, the stable-speed evaluation is
\begin{equation}
     r_v \triangleq 
\begin{cases} 
      \frac{v_{x}-\bar{v}_{\min}}{\bar{v}_{\text{target}}-\bar{v}_{\min}}, & \bar{v}_{\min} < v_{{x}} \leq \bar{v}_{\text{target}}; \\
      
      \frac{\bar{v}_{\max} - v_{{x}}}{\bar{v}_{\max}-\bar{v}_{\text{target}}}, & \bar{v}_{\text{target}} < v_{{x}} \leq \bar{v}_{\max};\\
      0, & v_{{x}}\leq \bar{v}_{\min}\;\, \text{or}\;\, v_{{x}}>\bar{v}_{\max}; \\
   \end{cases}
\end{equation}
where $v_{{x}}$ is the current longitudinal speed of the vehicle, constant $\bar{v}_{\min}$, $\bar{v}_{\text{target}}$ and $\bar{v}_{\max}$ are minimum, target and maximum speeds, respectively. 
Any speed larger than the maximum speed or less than the minimum speed is suppressed. Speeds $\bar{v}_{\min}$ and $\bar{v}_{\max}$ can be changed according to different traffic conditions. 
We want the ego vehicle to stay at the center of the road. 
So, the lane-centering evaluation is
\begin{equation}
    r_c \triangleq 
    \begin{cases}
          -1, & |y-y_c| \geq \bar{d}_c; \\
          0,  & \text{Otherwise}, 
    \end{cases}
\end{equation}
where $y_c$ is the lateral location of center of the current lane, and $\bar{d}_c$ is a constant distance threshold. 
Lastly, the ego vehicle should keep a safe time headway and distance. 
The headway evaluation is defined as
\begin{equation}
    r_h \triangleq 
    \begin{cases}
          -1, & \frac{d_{f*}}{|v_{f*}-v_x|}< \bar{T}_{\min} \;\, \text{or}\:\, d_{f*} < \bar{d}_s; \\
          0, & \text{Otherwise},
    \end{cases}
\end{equation}
where affordance indicators $d_{f*}$ are either $d_{fl}$ or $d_{fr}$ whichever shares the same lane; $v_{f*}$ is defined in the same way; and constants $\bar{T}_{\min}$ and $\bar{d}_s$ are safety thresholds.


      
      
      
   




\subsection{Safety Supervisor \& Low-level Controller}
\subsubsection{Safety Supervisor}
To train the ego vehicle safely and avoid frequent resets due to collisions, the safety supervisor makes use of the human lane-changing model.
Since the physical variables observed by the driver in a MV, $v_s$, $v_c$, $v_f$, $v_b$, $d$, can also be measured by the AV through its sensing system, the lane-changing decisions of the driver can be predicted by the AV through the human lane-changing model.
Whereas there are different types of drivers, in this work we assume all drivers are the same; we save the task of modelling and identifying types of various drivers as our future work.

Using these predictions, the safety supervisor can evaluate the consequences of actions from the RL agent.
Regarding inter-vehicle consequences, within a short prediction time horizon $t_{\text{pred}}$, 
the future locations of a MV is estimated based on the predicted lane-changing decision and the current velocity of the MV. 
Likewise, the future locations of the ego vehicle within $t_\text{pred}$ is also estimated based on its current action and velocity.
A collision is predicted if the distance between the MV and the ego vehicle is within a predefined threshold $\bar{d}_s$ at any moment according to the projected trajectories.
The action is labelled as unsafe.
The safety supervisor reselects and replaces action as follows (Lines 6-11 in Alg. \ref{safeRL}).

\begin{itemize}

\item If the unsafe action is to change lanes, then the replacing action is to stay in the current lane instead.

\item If the unsafe action is to speed up, and the ego vehicle will choose to slow down to avoid collisions.
\end{itemize}

Sometimes, the consequences involve only the ego vehicle, e.g., the ego vehicle chooses an action that pulls it off-road. 
In such cases, the safety supervisor predicts the trajectory and determines that the ego vehicle will be off-track.
Then, it provisions a default safe action, for instance, lane keeping.

The actions which are admitted by the safety supervisor are labelled as safe actions.
They are sent to the lower-level controller, which controls the AV to interact with the environment and generate rewards according to Eqn. (\ref{eqn:reward_function}).
A safe action, the states before and after the action, and the corresponding reward are considered as a safe experience.

To fully utilize the experiences,
an unsafe action and the associated experience is not simply discarded; we keep unsafe experiences by attaching appropriate penalties and recording the associated states.
We store the unsafe experiences along with the safe experiences into the experience replay buffer. 
Every time we sample a mini-batch experiences from the replay buffer, we use the experiences to update our policy (Lines 12-25 in Alg. \ref{safeRL}).

\subsubsection{Low-level Controller}
Once receiving an action from the high-level agent,
the low-level controller controls the vehicle directly. 
This hierarchical design greatly reduces the training time compared to methods using agents to output control signals directly. 
For a low-level controller, classical feedback control methods, for instance, PID and MPC, are good choices. 
In this work, we use PIDs for both lateral (steering angle) and longitudinal controls (throttle).

\subsection{The SafeRL Algorithm}

Our SafeRL algorithm is shown in Alg. \ref{safeRL}. 
We use an improved version of DQN called Double deep Q-Network (DDQN), which mitigates the over-estimation problem of DQN~\cite{van2016deep}. 
Though we used the DDQN in our experiment, the proposed framework is suitable for other RL algorithms.

As parameters, $M$ is the total number of training epochs, $T$ is the total training time in each epoch, and $K$ is the size of sampled experiences at each time.
After initialization, line 5 shows the action selection using $\epsilon-greedy$ method \cite{watkins1989learning}. 
Lines 6--11 illustrate how the safety supervisor works: 
Every time after the RL agent chooses an action $a_t$, the supervisor checks whether this action is safe or not. 
It is replaced by a safe action ${a_t}'$ if it is determined unsafe. 
Line 8 stores the unsafe experience $(s_t, a_t, *, r_{col})$ to the replay buffer $\dcal$, where $s_t$ is the previous state, $*$ means no next state because of collision, and $r_{col}$ is the penalty of collision.
After the agent takes the safe actions, Lines 12--17 save the corresponding experiences with different rewards, $r_{col}$ or $r_{t+1}$, to the replay buffer.
Lines 18--25 update the Q-network.
Line 18 samples a mini-batch of experiences from the replay buffer.
Lines 19--22 estimate the value of the policy by the target network either as $r_{t+1}$ or as
\begin{equation}\label{eqn:DDQN}
    Y^\text{DDQN}_{\text{target}} \triangleq r_{t+1} + \gamma Q_{\phi_N} (s_{t+1}, {\text{argmax}}_{a'} Q_{\phi} (s_{t+1}, a'))
\end{equation}
Lines 24 calculates the gradients with respect to ${\phi}$ (Eqn. (\ref{eqn:update_phi})) and  updates the Q-network. 
Line 25 updates the target network every $N$ steps and keeps it fixed at other steps.

\begin{algorithm}[H]
\caption{SafeRL for autonomous driving} \label{safeRL}
\begin{algorithmic}[1]
\renewcommand{\algorithmicrequire}{\textbf{Parameters:}}
\renewcommand{\algorithmicensure}{\textbf{Output:}}
\REQUIRE{ $M, T, K$}
\STATE{Initialize the Q-network, $Q_{\phi}$; the corresponding target network $Q_{{\phi}_N} \leftarrow Q_{\phi}$; and the safe replay buffer $\dcal \leftarrow \emptyset$\;}
\FOR{$j = 0$ to $M-1$}
	\STATE{ Initialize $t \leftarrow 0$ and initial state $s(0) \leftarrow s_0$}
	\WHILE{$t<T$}
	    \STATE{Select a random action with probability $\epsilon$, otherwise select action $a_t \leftarrow {\text{argmax}}_{{a}'} \hspace{0.05 cm} Q_{\phi}(s_t, {a}')$}
	    \IF{$a_t$ is unsafe}
	        \STATE{ Replace it with a safe action ${a_t}'$}
	        \STATE{Store ($s_t$, $a_t$, *, $r_{col}$) to $ \dcal$}
        \ELSE
            \STATE{${a_t}'\leftarrow a_t$}
        \ENDIF
        \STATE{ Perform ${a_t}'$ and observe ${s_{t+1}}$, $r_{t+1}$}
        \IF{termination}
            \STATE{ Store ($s_t$, $a_t$, *, $r_{col}$)$ \text{ to } \dcal$}
        \ELSE
            \STATE{ Store ($s_t$, $a_t$, ${s_{t+1}}$, $r_{t+1}$)  $\text{ to } \dcal$}
	        \ENDIF
	   \STATE{ Sample a mini-batch of size $K$ from $\dcal$}
	   \IF{termination}
	        \STATE{ $ Y^\text{DDQN}_{\text{target}} \leftarrow r_{t+1}$}
	   \ELSE
	        \STATE{Update $Y^\text{DDQN}_{\text{target}}$ according to Eqn. (\ref{eqn:DDQN})}
	   \ENDIF
	   \STATE{Update $Q_{\phi}$ according to Eqn. (\ref{eqn:update_phi})}
	    \STATE{ Update $Q_{{\phi}_N} \leftarrow Q_{\phi}$ every $N$ steps} 
	\ENDWHILE
\ENDFOR
\end{algorithmic}
\end{algorithm}

\section{Experiment, Results, \& Discussion}\label{sec:6}
In this section, we will present the experiments conducted for identifying the parameters of the proposed human lane-changing decision model, as well as for evaluating our SafeRL methods. The  experiments are performed on an open-source driving simulation platform CARLA~\cite{Dosovitskiy17}.

\subsection{Variables \& Parameters in Human Lane-changing Model}
To exploit the human drivers' regret decision model in AV learning, values of the objective variables, $v_s$, $v_c$, $v_f$, $v_b$, $d$, and the parameters, $\sigma_1$, $\sigma_2$, $\sigma_3$, $\eta_1$, $\beta_1$, $\beta_2$, $\tau_s$, must be obtained. 
We assume that human drivers are well informed about the driving-related objective variables, which are inputs to the decision-making model in Eqn. (\ref{eqn:driver_decision_model}). 
On the other hand, the parameters are driver-specific constants that can be estimated through well-designed experiments.

\begin{figure}[!ht]
  \centering
\includegraphics[width=0.7\linewidth]{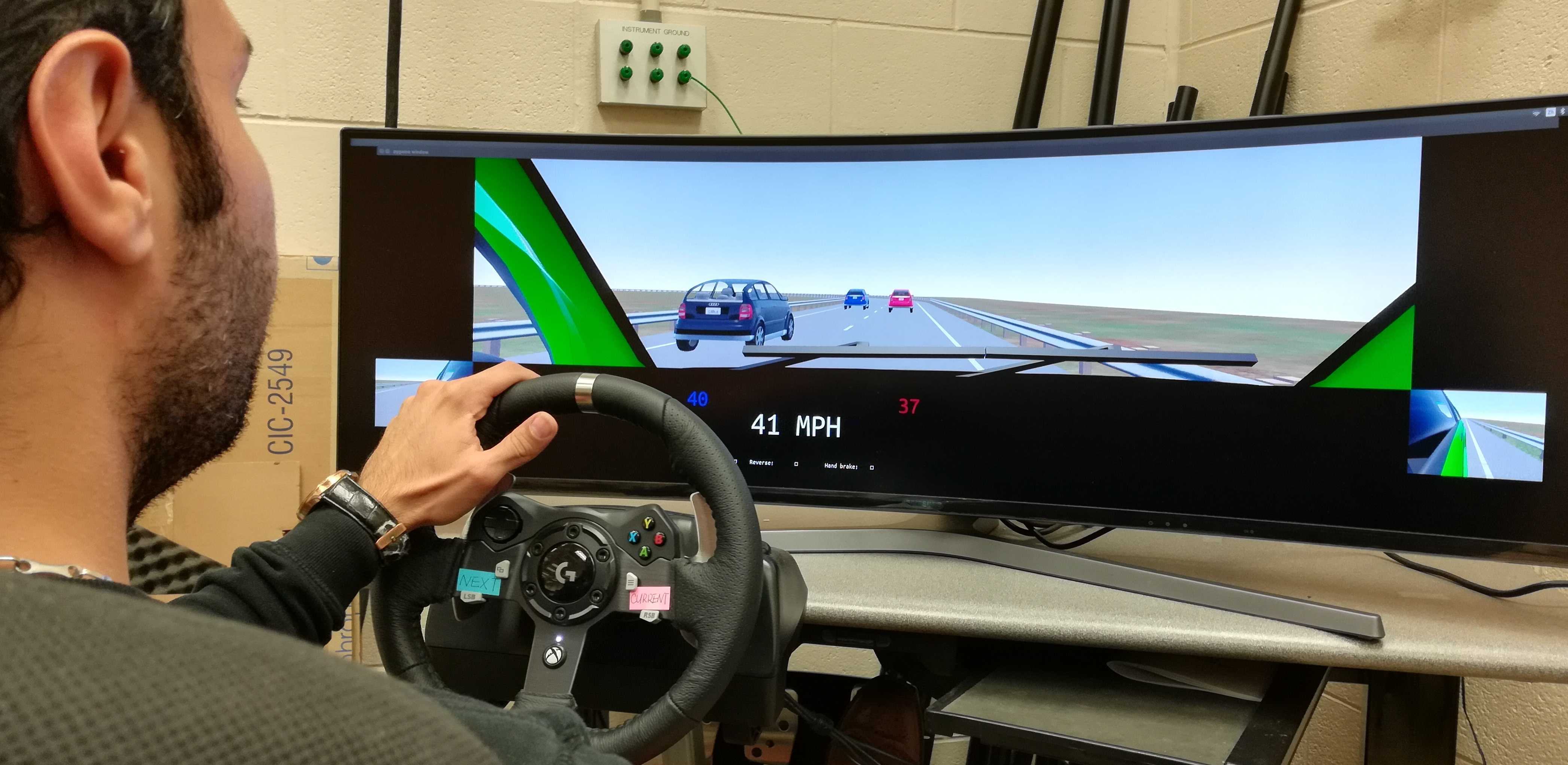}  
  \caption{A subject is on the simulator for collecting lane-changing decisions.}
  \label{fig:HumanDriver}
\end{figure}

Due to space limit, we will only sketch the procedure of parameter estimation through experiments in this work. We developed a driving simulator based on CARLA as in Fig. \ref{fig:HumanDriver}. As a pilot study, one subject was invited to drive in a two-lane traffic scenario. 
The environment vehicles were set up with different speeds and distances to test the subject's lane-changing decision-making. 
The objective variables, $v_s$, $v_c$, $v_f$, $v_b$, $d$, and the lane-changing decisions, option C or option K, were collected to construct a labelled data set. 
We then used logistic regression to fit the parameters to the dataset. 
The accuracy of the fitted model on the data set is $83.33\%$.
Tab.~\ref{tb:HM_params} summarizes the parameters we obtained.
We have already applied IRB for this project and will perform more human subject tests in our future work.
\begin{table}[!h]
\renewcommand{\arraystretch}{1.3}
\caption{Parameters for human lane-changing decision model.}
\label{tb:HM_params}
\centering
\begin{tabular}{c|c|c|c}
\hline
 $\sigma_1$ & $\sigma_2$ & $\sigma_3$ & $\eta_1\; (\text{m}^2/\text{s}^2)$      \\    \hline
  10.1795 & 0.1130 & 0.5108 & $152.5796$\\    \hline
 $\beta_1$ & $\beta_2$ & $\tau_s \; (\text{s})$  &  \\    \hline
 $9.9170$  & $2.3812$ &  $3.5193$  &\\\hline
\end{tabular}
\end{table}

\subsection{Experimentation of SafeRL}

We created a 400-meter standard two-lane road scenario in CARLA.
The driving scenario setup is shown in Fig. \ref{fig:TrafficScenario}, where the blue one is the AV (ego) while the green and the red one are two MVs.
The current speed $v_c$ of the red MV is 5.56 m/s and its best speed $v_b$ is also 5.56 m/s.
On the other hand, the speed $v_c$ of the green MV is 5.56 m/s while its $v_b$ is 12.5 m/s. 
Since the green MV is behind the red, it may want to change lanes.
The distance  between the blue ego vehicle and the green MV, $d$, is 10 m.
The MVs are assumed to use the regret-based human lane-changing model identified above for high-level decision-making. Low-level controls of MVs are implemented uisng the default PID controllers in CARLA.
The ego vehicle was controlled by the SafeRL agent.
The dynamical models of all the vehicles were provided by CARLA. 

The trajectories of both the AV and MVs are predicted using the Euler's method:
\begin{subequations}
\begin{IEEEeqnarray}{rcr}\label{4}
v_x(t + 1) &\,=\,& v_x(t) + a_x(t) \Delta t,\label{eqn:desired_traj_a}\\
x(t+1) &=& x(t) + v_x(t) \Delta t, \label{eqn:desired_traj_b}\\
y(t+1) &\,=\,& v_y(t) + v_y(t) \Delta t\label{eqn:desired_traj_c}.
\end{IEEEeqnarray}  
\end{subequations}
For the AV, the longitudinal acceleration $a_x$ and lateral velocity $v_y$ are from the RL agent as discussed in Sec. \ref{sec:action_space}. The safeRL agent was trained in CARLA simulations for 1500 epochs. 
At the beginning of each training epoch, all the vehicles started from prespecified positions. An epoch will stop when a collision happens or the ego vehicle reaches
the end of the road.


The Q-network $Q_\phi$ and the target network $Q_{\phi_N}$ are neural networks of 2 fully-connected layers and each layer has 64 nodes followed by ReLU activation. 
The networks are trained by Adam optimizer \cite{kingma2014adam} with learning rate $\alpha = 1e-4$. 
The exploration rate is continuously annealed from 1 to 0.05 over the first 1000 epochs and then kept constant for the remaining epochs. 
Actions are updated every two steps \cite{li2017explicit}.
Parameters for the reward function, network training, and Euler's equation are shown in Tab.~\ref{tab:params}.
\begin{table}[!h]
\renewcommand{\arraystretch}{1.3}
\caption{Simulation parameters.}
\label{tab:params}
\centering
\begin{tabular}{c|c|c|c|c|c}
\hline
 $w_s$ & $w_v$ & $w_c$ & $w_h$  & $\bar{d}_c$ & $\bar{d}_s$ \\    \hline
2000 &10 & 3 & 15 & 0.5 m & 18 m\\   \hline
 $\bar{T}_{\min}$ & $\bar{v}_{\text{target}}$ &$\bar{v}_{\min}$ &  $\bar{v}_{\max}$  & $t_{\text{pred}}$ & $M$ \\    \hline
2 s & 12.5 m/s  & 5.56 m/s &16.67 m/s  & 0.7 s & 1200\\    \hline
 $K$ & $\gamma$ & $\Delta t$ & $\bar{a}_x$ &  $\bar{v}_y$ &\\ \hline
256  & 0.99 & 0.1 s &  2 m/s$^2$  &  1.8 m/s  & \\\hline
\end{tabular}
\end{table}

\subsection{Results and Discussion}
\begin{figure}[!ht]
  \centering
  \includegraphics[width=0.35\textwidth]{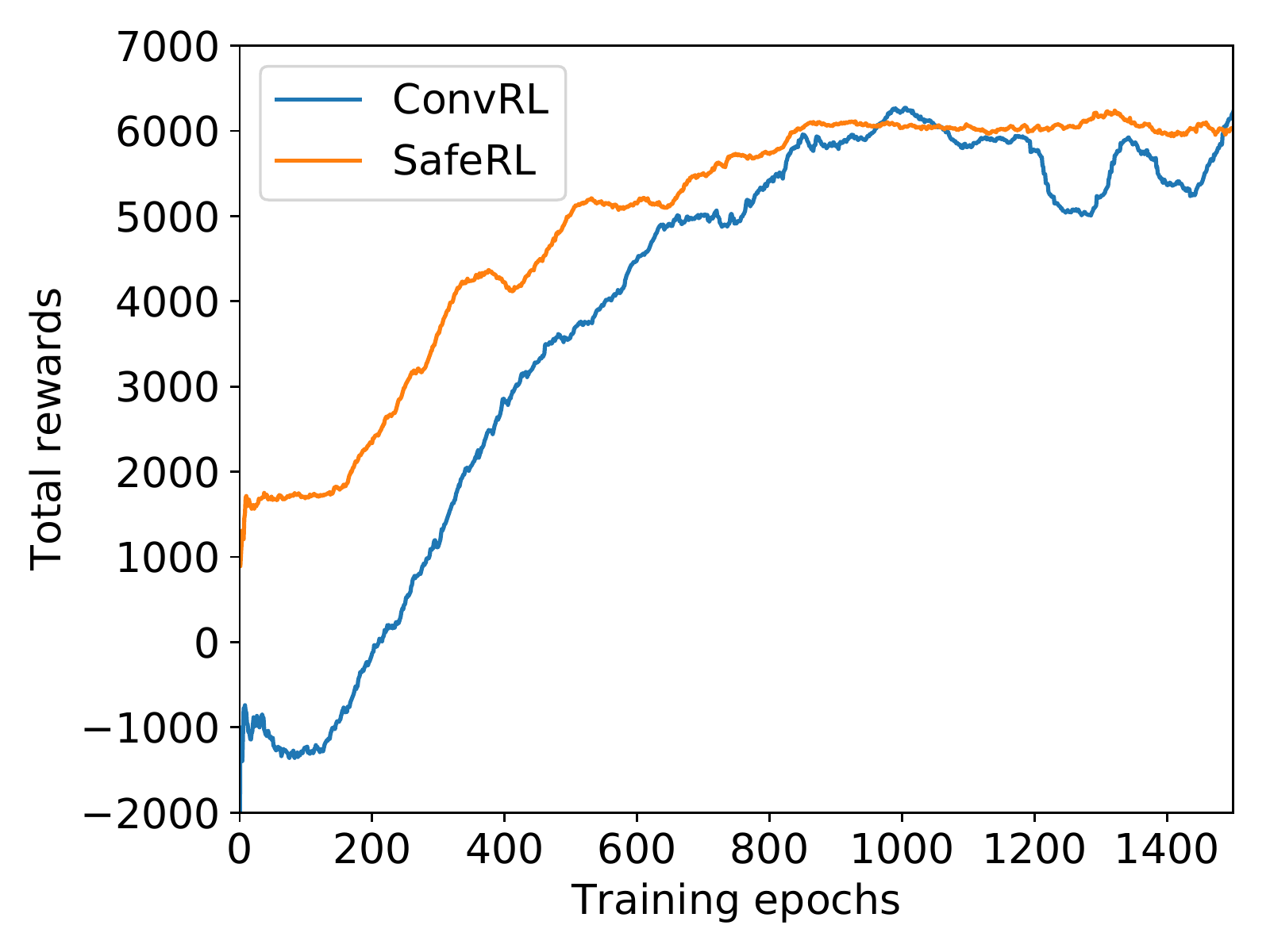}
  \caption{Learning curves of SafeRL and ConvRL}\label{train_return}
\end{figure}

We compare our proposed SafeRL algorithm against the conventional DDQN without safety supervision (we call it ConvRL thereafter).
Fig. \ref{train_return} shows the learning curves of the SafeRL and the ConvRL. 

It is clear that SafeRL quickly leads to a reasonably good control policy with much larger rewards than ConvRL in the first 200 epochs. 
This is because the SafeRL  is able to avoid collisions and thus offers more training experience without early epoch termination. 
Indeed, Tab. \ref{tb:collisions} shows that there was no collisions during training when using SafeRL, whereas using the ConvRL 8.6\% epochs ended up in collision. 
Safety is guaranteed during training when using the SafeRL.

We should note that the safety supervisor in the SafeRL decreases the exploration space to avoid collisions.
As a result, the available explorations of the SafeRL are less than the ConvRL.
However, in Fig. \ref{train_return}, as the training progresses, the two curves converge almost at the same level, around the 800th epoch, despite the fact that the SafeRL has a smaller exploration space. 
This is reasonable because the optimal policy for choosing actions should be within the constraints set by the safety supervisor. 
The action space that SafeRL cannot explore is the part that the AV should indeed avoid.  
\begin{table}[!th]
\renewcommand{\arraystretch}{1.3}
\centering
\caption{Collision Comparison.}
\label{tb:collisions}
\begin{tabular}{c|c|c}
\hline
                & Collision amount & Collision Ratio  \\ \hline
    ConvRL & 129    &  8.6\%  \\ \hline 
    SafeRL & 0 & 0 \\ \hline

\end{tabular}
\end{table}

\begin{figure}[!ht]
  \centering
  \includegraphics[width=0.35\textwidth]{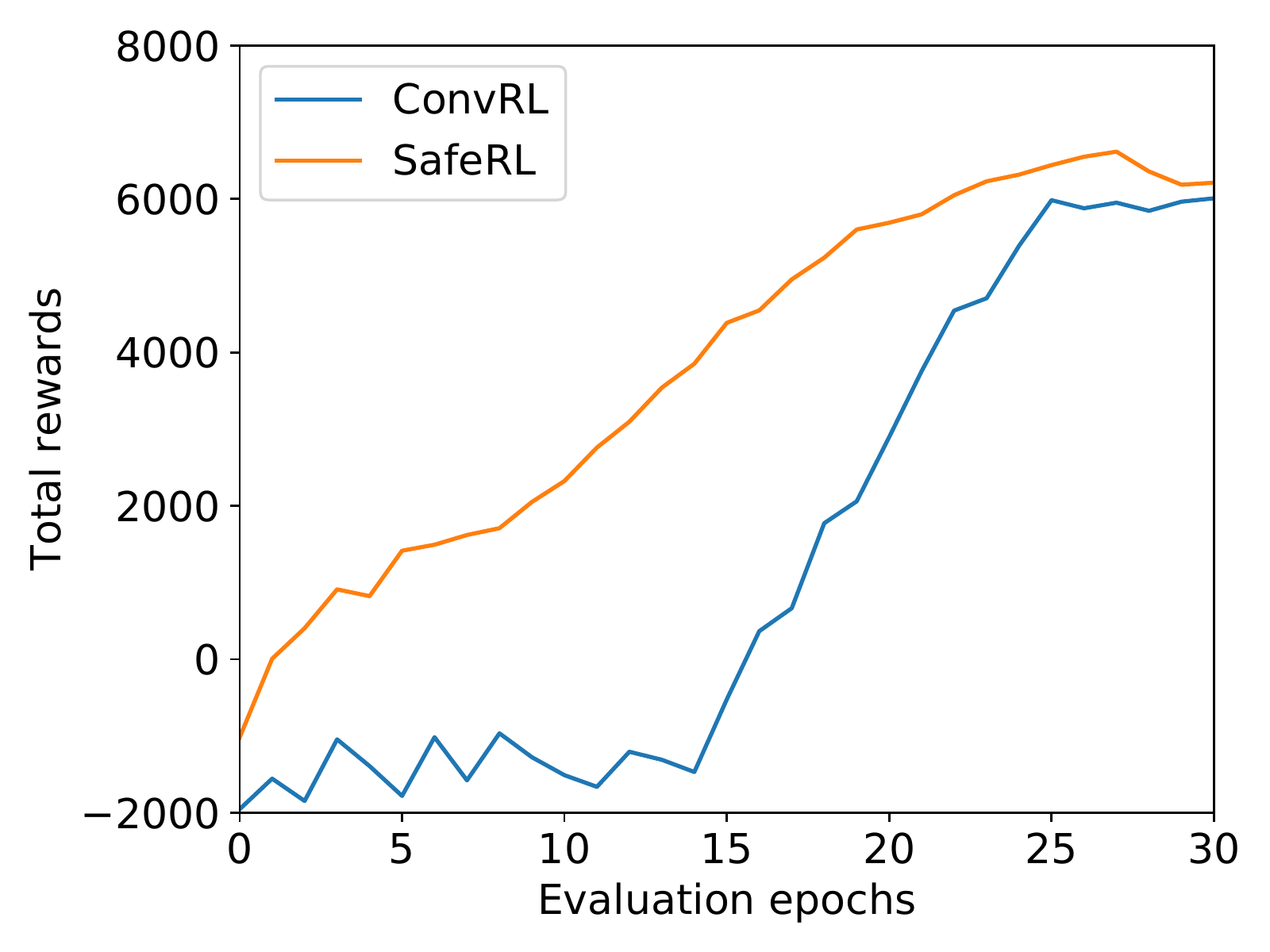}
  \caption{Evaluating curve of SafeRL and ConvRL.}\label{eval_return}
\end{figure}

Conversely, the unconstrained exploration by the ConvRL leads to slightly degraded evaluation performance. 
Fig. \ref{eval_return} shows performance of the policies evaluated every 50 epochs during training. 
The total rewards of an epoch of the SafeRL  are always better.
Its learning was smoother and more stable as evidenced by the constant improving rate. 
The initial setbacks of the ConvRL were due to the many collisions. 

After both the models converge, after the 800th epoch in Fig. \ref{train_return}, the SafeRL is more stable. 
The ConvRL still has large fluctuations even after convergence. 
This is because collisions still happen. 
So even the completed trained ConvRL model still cannot guarantee to be collision-free. 
The SafeRL, nevertheless, ensures no collision for the trained model.  
We hence demonstrate the efficacy of the proposed SafeRL with safety guarantee as well as more efficient and stable learning.



\section{Conclusions}\label{sec:7}
We presented a framework for RL to incorporate a safety supervisor which uses human lane-changing decision model for making predictions.
We developed the human lane-changing model and did pilot testing to show its validity.
The predictions help RL learn its driving policy safely and stably. 
Experimental results showed that our proposed SafeRL can reduce collisions to zero during training and implementation, while keep the training performance not impinged. 

As the future work, we will make our method more robust to different traffic scenarios and further improve the learning efficiency for faster learning rate.

\bibliographystyle{ieeetr}

\end{document}